\documentclass[12pt]{drpaper}

\title{Language and Thought:\\The view from LLMs}
\author{Daniel Rothschild\\draft\\forthcoming in Lepore and Sosa (eds.) \textit{Oxford Studies in Philosophy Language}}

\newcommand\blfootnote[1]{%
  \begingroup
  \renewcommand\thefootnote{}\footnote{#1}%
  \addtocounter{footnote}{-1}%
  \endgroup
}
\begin{document}

\maketitle

\blfootnote{I am extremely grateful to Nick Shea, Matthew Mandelkern, and Emmanuel Chemla for extensive comments on earlier drafts.  Thanks also to William Ratoff for an important critical comment at a presentation of this paper.}

\section{The Cognitive Utility of Language: A Lesson from the Great AI Experiment}

Recent work in AI constitutes one of the most expensive and ambitious scientific experiments in human history. By designing and building massive artificial neural networks  with a variety of different architectures and training protocols, the performance characteristics of this form of AI is finally emerging. The \emph{Great AI Experiment} is happening now.

In what sense is the development of AI a scientific experiment? In cheaper scientific ventures like the CERN particle collider (c. \$4.75 billion) or the Hubble Space Telescope (c. \$16 billion) it was relatively clear what the scientists aimed to test or discover. The nature of subatomic particles! The history of the universe! When engineers and scientists build and test and AI system though what are they aiming to find out? Why is AI development any more a scientific venture than the latest developments in electronic vehicles or heat pumps? Isn't it all just engineering and technology?

The Great AI Experiment is an  experiment in cognitive science. By building machines capable of performing the kind of tasks routine for human minds (and sometimes animal ones), we are collecting indirect evidence about how biological minds work. Given how little we understand about biological minds, our own or other animal's, even such indirect evidence is immensely valuable. 

Another reason it is hard to see the Great AI Experiment as a scientific endeavor is that the results coming out of it, are generally, in philosophical parlance, \emph{a priori}. How a machine with a given structure will behave when trained in a certain way is not contingent the way the nature of sub-atomic particles is. Do enough calculations and you will discover what any particular AI system you can build will do. Nonetheless, practically speaking the only way to find out how today's complex AI systems will work is to build them.\footnote{For example, high-performing Large Language Models (LLMs) today have billions of parameters and are trained on trillions of tokens \citep{touvron2023llama}.} Building a simulation of an AI system to find out how it will operate is no different than building the system itself.

Like any good scientific experiment the results have been surprising. Perhaps none more so than the performance of Large Language Models (LLMs), from the early LLMs such as Google's Bert \citep{devlin2018bert} to more recent juggernauts like OpenAI's GPT-4 \citep{devlin2018bert}. The performance of LLMs is surprising in at least two respects. First, LLMs are a massive leap forward in text-based AI. LLMs write original verse, converse cogently on almost any topic, and are able to process and fairly intelligently summarize large complex texts. Few guessed in the mid-2010s this level of AI performance was on the horizon.

Perhaps even more surprising is the basic computational mechanisms that leads to such successes. While the inner workings of commercial LLMs, like OpenAI's GPT-4, are trade secrets, the core computational architecture and training regime behind  them is in the public domain. At any LLM's heart is a massive transformer network  trained to perform \emph{next token prediction} on texts \citep{transformer,devlin2018bert}. That is, LLMs are trained, given a sequence of words as a prompt, such as ``The quick brown fox jumped over the lazy \rule{1cm}{0.15mm}'', to predict the missing word (``dog''?). LLMs generate text by predicting, word by word, how a text will continue.\footnote{Natural language words are broken into smaller parts, \emph{tokens}, but the details of this are not relevant here. Under most tokenizations used today there are about 1-3 tokens per word, so the number of words is at the same order of magnitude as the number of tokens.} 

Before the creation and deployment of LLMs it was hard to imagine a system performing next-token prediction that could write and reason as well as LLMs do. Of course, there was no doubt that a system trained to do next-token prediction would produce output. It might have  been expected that the system, trained on enough data, would master some grammatical rules of English and be able to stay roughly on topic as it generated new texts to continue old ones. Connectionist networks have long been acknowledged to be very good at spotting and imitating patterns, and with enough computation power and training it is not surprising that they would capture many of the more basic patterns in large stores of texts. But that such systems---even augmented in various other ways---would have the power to make up jokes, to engage in complex reasoning, to mimic styles was a shock to those both in and out of  AI.\footnote{A dramatic public display of surprise was Geoffrey Hinton's when he quit Google in order to publicly voice his apocalyptic worries about progress in AI \citep{HintonBatshit}.} This is an example where an outcome of the Great AI Experiment, like those of many great scientific experiments, was surprising and informative.

My focus here is what the scientific takeaways so far are from the Great AI Experiment? The most obvious one is a 
 demonstration of the power 
 of the system of computational paradigm variously called artificial neural networks (ANNs), connectionism, parallel distributed processing (PDP) or, most recently, deep learning.\footnote{There are different shades of meaning for these terms. Networks in deep learning, for instance, for instance tends to have more hidden layers (often thousands rather than just a few) than traditional PDP networks \citep{deeplearninglecun}.}
Such systems have long been put forward by some cognitive scientists as a computational model for animal and human reasoning \citep{10.5555/104279}. Connectionism is a self-styled \emph{subsymbolic} approach to AI, providing an alternative to traditional AI where cognition is assumed to take place using a symbolic computational architecture analogous to that used in standard computers. The top performing AI at the turn of the century were mostly traditional symbolic AI. While connectionism had many significant accomplishments under their belt by this time---including in both visual recognition capacity and board game play---it was still possible (and indeed common) to dismiss it as an extremely limited computational paradigm.\footnote{\citet{Carruthers2002-CARTCF-2}, for example, wrote, `` The successes of the distributed connectionist program have been limited \ldots mostly being confined to various forms of pattern recognition; and there are principled reasons for thinking that such models cannot explain the kinds of structured thinking and one-shot learning of which humans and other animals are manifestly capable.'' \citet{Samuels2010-SAMCCA} similarly wrote ``\ldots connectionist accounts of reasoning are extraordinarily thin on the ground; and at the time of writing, there are no remotely plausible connectionist accounts of human reasoning.'' Connectionism can no longer be so dismissed.} Now across AI for tasks such tasks as video game play, board game play, computer vision, and voice recognition, ANNs perform at a level far above that of traditional symbolic AI.

The take-home for cognitive science is inevitable: if artificial neural networks are behind the greatest advances in AI in the last half-century then perhaps our own brains also use some of this technology \citep{Buckner2023-BUCFDL}. The natural conclusion is buttressed by the many connections between actual brain structure and the structure of ANNs, that have long inspired those in the field.\footnote{These parallels go well beyond the imperfect resemblance between nodes in ANNs and neurons in the brain. For examples, convolutional neural networks (CNNs) which are used for applications such as image, video, and sound processing are designed with a very similar structure to parts of the visual cortex.} Of course, our human brains look and act vastly different from any ANN around today, but the successes of  ANNs combined with the parallels to biological brains suggest we have much to learn from them. 

Despite the vindication of the subsymbolic architectures through the Great AI Experiment, symbolic computation may yet to have an important role to play in human thought. Before the Great AI Experiment many scientists and philosophers argued that artificial neural networks needed to be supplemented  with a classical computational architecture to achieve human-level reasoning \citep{Pinker2002-PINTPD-2,bookmarcus2003algebraictitleThealgebraicmind}. Others proposed that subsymbolic computational systems may realize symbolic architectures at a higher level \citep{smolensky1988proper}. Descendants of these positions are still are very much alive.\footnote{\citet{marcus2018deep, marcus2022deep} takes an extreme position here, \citet{Lake2017-LAKBMT} and \citet{Quilty-Dunn2023-QUITBG, SmolenskyMcCoy} are more cautious defenders of a role for symbolic architecture.} The fact that many successful AI systems  supplement their neural networks with classic computational systems can be seen to as evidence for the continued relevance of symbolic processing in AI.\footnote{For example, Google DeepMind's Go playing systems like AlphaZero \citep{doi:10.1126/science.aar6404} rely on a classical stochastic search system to supplement the connectionist network.} The debates of late 20$^\text{th}$ century cognitive science continue, albeit with a playing field now heavily stacked in favor of the subsymbolic where it once might have been the reverse. 

I will argue here that the successes of LLMs, while vindicating connectionist approaches to computational architecture, are \emph{also} a demonstration of the importance of symbolic reasoning in AI, in particular the symbolic system that is natural language.
LLMs are uniquely good at domain-general inference---they have a breadth of reasoning power unmatched in AI today. This unique success, I argue, evidences the power of natural language itself as a medium of thought. So, while the Great AI Experiment overall shows the power of \emph{subsymbolic} architecture, LLMs points to a central role in cognition for the \emph{symbolic} system of representation that is natural language. 

This result  has profound implications for one of the biggest questions in cognitive science: What is the role of language in thought? In 1996 Daniel Dennett speculated:
\begin{quote}
Perhaps the kind of mind you get when you add language to it is so different from the kind of mind you can have without language that calling them both minds is a mistake.
\end{quote} Others, like \citet{FodorLOT,BloomKeil} and \citet{fedorenko2024language} argue that the cognitive utility of language may be much smaller.
The development and deployment of LLMs can be seen as a large-scale test of the effect of language on thought. All in all,  the results  support Dennett's radical thesis that ``adding language'' to a cognitive system has a transformative effect on it. The giant leap forward that LLMs have made for AI in common sense and general reasoning is only possibly because of their use of natural language.

Reflecting on the Great AI Experiment can also give us some insight into \emph{why} natural language makes such a difference in AI. 
 The performance profile of recent AI systems indicates that it is the  abstract, data-efficient nature of linguistic representation that accounts for its cognitive utility. In a slogan, \emph{Natural language makes general inference computationally tractable.}

\section{The argument}

In this section I will rehearse the core argument of the paper, reserving detailed argument for later.

There are three main parts. The first part is a presentation of an important \emph{result} from the great AI experiment: \emph{only AI systems with access to natural language perform well at general reasoning}. The second part is an  \emph{explanation} for that result: \emph{the abstraction and compression in natural language makes general reasoning computational tractable}. The third part assess the implications of these results in AI for biological minds like our own.

\noindent \emph{The result}

The result from the great AI experiment I want to highlight is that only AI systems extensively trained with  natural language show powerful domain-general reasoning. This is both a positive and negative claim: The positive side that  what I call \emph{language-based systems}, such as LLMs, exhibit powerful domain-general reasoning, the negative side is that AI-systems without extensive language training, non-language-based systems, do not.

Once we get straight on a reasonable notion of what domain general reasoning is, I don't think either claim should be very controversial. I take the notion of domain from cognitive psychology, where types of cognitive capacities are divided up according to subject-matter.\footnote{The notion of domain is used by Fodor in his classic discussion of modularity \citeyearpar{FodorModularity}. See also \citet{spelke2022babies} and \citet{Carey2009-CARTOO-3}.} There is the domain of agency, our reasoning about other animals with goals and desires, the domain of location, where we reason about our location in the world, the domain of number, our reasoning about numerosities, and so on. A cognitive system, like a human mind, engages in domain-general reasoning when it is able to reason across a wide swath of these domains, integrating information from different domains into a single inference. (This is, of course, a largely anthropocentric notion as  generality here is just generality across those domains humans reason within.)

I argue in the next section (\S\ref{onlygameintown}) that high-performing LLMs, like GPT-4 or Claude, engage in powerful domain general reasoning. I am not arguing that LLMs  exhibit something close to Artificial General Intelligence (AGI) understood as a super-human level of reasoning. LLMs have notable limitations, particularly  with respect to the kind of precise, careful reasoning needed for logic, mathematics, and maintaining the thread of  long arguments more generally \citep[see, e.g.\,][]{MAHOWALD2024517,dziri2024faith}. Despite such limitations, LLMs are still  powerful engines of general inference, outpacing any other form of AI in this task and, in limited respects, individual human minds.

I argue as well, that there are no comparable capacities of general inference in ANNs that do not rely extensively on natural language. Moreover, this lack is not a simple artifact of the fact that networks without language are often designed and trained for fairly limited purposes (e.g. facial or speech recognition or playing board games). Rather the results of the  Great AI Experiment show that domain-general and common-sense reasoning is, for now, beyond the capacities of non-language-based AI systems.

\noindent \emph{The explanation}

My explanation, in section \ref{utilityoflanguage}, of this result is that \emph{LLMs are powerful reasoners because natural language as a medium is compressed enough to make reasoning computationally tractable}.

A comparison with video is illustrative. A 30-second-video at the quality of a very low quality video call requires about 2.5 megabytes to store. Tolstoy's entire \textit{War and Peace}, by contrast takes up about 3 megabytes. The extraordinary compression and abstraction of information in natural language is often argued to be a property of communicative utility, but it is also a crucial factor in the utility of language for inference. The more options you have the harder inference and prediction are. Natural language, by allowing lean descriptions of situations, encoding exactly the facts that we care about, at the expense of other details, makes inference and prediction tractable.  It is perhaps only with such compresed linguistic encoding that we can even feed an AI system enough relevant data to train it to preform general inference.

Next token prediction---with the powerful computers and massive training data behind behind LLMs---gets some purchase with natural language, because of how efficient it is. These same massive computational powers when trying to perform general inference in a realm that is not as abstract, such as video, are comparatively helpless. They struggle to create  sensible, let alone the probable continuations of video in a manner analogous to next token prediction. 

The Great AI Experiment  points to the utility of natural language as a medium of thought---at least for the current generation of AI systems. 

\noindent\emph{Implications for cognitive science}

In the final section (\S\ref{lessons}), I discuss implications of these conclusions about AI systems for our biological minds. I argue, first, that the success of LLMs makes plausible the idea that natural language is optimized, at least partly, for its utility in thought, not just its communicative utility. This goes against a  common view in cognitive science that language is shaped soley for communication. More generally, the utility of natural language for thought supports a robust role for language in thought.

The success of LLMs also can help shape the debate over the effect of language and thought. LLMs provide a possibility proof that learning a language---by exposure to large amounts of text---can alone unlock significant cognitive abilities. However, for the case of us humans, it remains an open question to what degree  our own cognitive abilities are shaped by language, rather than language being shaped by our pre-linguistic cognitive abilities. The success of LLMs complements other evidence of the power of language to shape cognition.

\section{LLMs are the only artificial general reasoners in this town}\label{onlygameintown}

The main point I want to make here should be uncontroversial. It’s that today’s Large Language Models, such as GPT-4, are the only AI systems that can be said to have substantial success at domain general reasoning. There is both a positive and a negative side to this claim.  The positive one is that  large language models  succeed at general inference. The negative side is that AI systems \emph{other than} large language models \emph{don’t} succeed at domain general reasoning. 

The reason the positive claim should be uncontroversial is partly that I’m not setting the bar too high. I am not claiming that the best LLMs systems are on par with average human performance across the bar. While they can do things that no single person can, LLMs also show  systematic and unsystematic limitations that people do not \citep[e.g]p[{MAHOWALD2024517}.  By succeeding at domain-general reasoning, all I mean is that a) these systems show signs of regularly and consistently making reasonably good inferences in problems across a wide range of the domains that people can reason within (e.g. language, causation, agency, logic, mathematics) and b) they do so in way that goes beyond pure memorization or its near equivalent.

That the performance of LLMs  goes  beyond regurgitation of training data is well established \citep[see][for a recent review]{milliere2024philosophical}. It is true that by training LLMs on effectively all publicly available texts and more, LLMs  function on the basis of orders of magnitude more linguistically-packaged information than any single person gets in a lifetime. Nonetheless, it's easily shown that LLMs create original texts in response to questions \citep{mccoy2023much}. In fact, even if LLMs answer a query by regurgitating a text in their training data, choosing what text to produce is itself an extraordinarily difficult inference problem. The only case in which LLMs are not, in effect, solving reasoning problems is the one in which the prompt and text produced are both in the training data as a single sequence. This kind of copying is not, generally, what LLMs are doing.\footnote{In fact, there is some direct evidence that LLMs work by building up models of the world \citep{li2022emergent,nanda2023emergent}.} There isn't enough training data in the world to allow this form of regurgitation.

Once we see that LLMs are not simply parroting their training data, the conclusion that they are able to perform inference needs little argument. We largely use LLMs exactly for their reasoning powers: their ability to summarize texts, to make reasonably intelligent conversation, and  to write original, cogent essays on almost any topic. More generally they are good at inference: describe a situation to an LLM and ask them to draw conclusions and they do a reasonable job across a wide range of domains. Such performance has been extensively tested and documented.\footnote{See, for example, \citet{kiciman2023causal},  \citet{bubeck2023sparks}, \citet{strachan2024testing}, and \citet{minaee2024large}.} 

Such facility with language was, for thinkers as diverse as Descartes and Turing, a critical mark of what counts as reasoning. In \emph{Discourse on Method} (1637) Descartes argued that the ability to intelligently converse was one of the marks that distinguishes true cognition, than what he thought possible from a purely mechanical device:
\begin{quote}
I made special efforts to show that if any such machines had the organs and outward
         shape of a monkey or of some other animal that lacks reason, we should have no means
         of knowing that they did not possess entirely the same nature as these animals; whereas
         if any such machines bore a resemblance to our bodies and imitated our actions as
         closely as possible for all practical purposes, we should still have two very certain
         means of recognizing that they were not real men. The first is that they could never
         use words, or put together other signs, as we do in order to declare our thoughts
         to others. For we can certainly conceive of a machine so constructed that it utters
         words, and even utters words which correspond to bodily actions causing a change in
         its organs [\ldots]. But it is not conceivable
         that such a machine should produce different arrangements of words so as to give an
         appropriately meaningful answer to whatever is said.
\end{quote}
Similarly  with his famous Imitation Game (the Turing Test), \citet{turingfamous} suggested we replace the question of whether machines can ``think'' with the question of whether they can respond in a human-like way to text-based conversation. For both seminal thinkers on minds and machines, the ability to converse intelligently was a mark of genuine rational capacities.

I have highlighted the general abilities of LLMs at conversation and commonsense reasoning. Let me hone-in though, in a specific way in which they show strong inferential abilities about a number of topics. We can feed an LLM as a prompt a description of a situation and ask (or otherwise induce) the LLM to tell us what it inferences it can draw about what will happen or what more facts obtain. Here is a typical example:\footnote{Created 7 August 2024 on GPT-4, this was my first try at creating an example of this sort. This is not presented as experimental evidence but only to give an example of the kind of skill explored in empirical work on the reasoning capacities of LLMs.}
\begin{quote}
\textit{DR}: Consider this scenario:\\
Suppose you put an unopened roll of toilet paper on a table that is uneven because 1 leg is 1 inch shorter than the other 3. What will happen to the roll of toilet paper?

\textit{ChatGPT-4o}:
If you place an unopened roll of toilet paper on a table with one leg that is 1 inch shorter than the other three, the table will be uneven and tilted towards the leg that is shorter. Here's what will likely happen to the roll of toilet paper:

\textbf{Roll of Toilet Paper Placement}: When you place the roll of toilet paper on the tilted table, gravity will cause it to move towards the lower side of the table.\\
\textbf{Movement of the Roll}: The roll of toilet paper will start to roll down the slope created by the uneven table. Depending on the angle of the tilt and the friction between the roll and the table surface, it might roll slowly or quickly.\\
\textbf{Stopping Point}: The roll of toilet paper will continue to roll until it encounters a barrier, such as the edge of the table, an obstacle on the table, or a point where the friction and the slope balance each other out, preventing further movement.\\
\textbf{Possible Falling}: If there is nothing to stop the roll and the tilt is significant enough, the roll of toilet paper might eventually roll off the edge of the table and fall to the ground.

In summary, on an uneven table with one leg shorter than the others, the unopened roll of toilet paper will roll towards the lower side and potentially off the table if no obstacles stop it.
\end{quote}
The answer involves inferences about tables, toilet rolls, and basic physics. It is not a particularly sophisticated piece of reasoning, but it is cogent. It's also the kind of inference people regularly need to make: non-deductive inferences about how situations will unfold. The ability of LLMs to make reasonable deductive and non-deductive inferences across a huge variety of domains, social, political, physical and mathematical, is perhaps the most surprising result of the Great AI Experiment.\footnote{It is worth (again) noting that the inferential powers of GPT-4 are likely due not solely to the next-token prediction using a transformer network but further, as yet non-public computational architecture in the system.}

Let me first respond to an obvious worry: Do LLMs really display \emph{domain-general} reasoning? Aren’t LLMs, in a sense, working in a single domain, namely by mastering language?
After all, most LLMs don’t do anything but create texts, and those that can do more don't necessarily thereby, show more powerful general reasoning capacities. However, that LLMs work in language doesn’t mean that all they do is manipulate language.  An AI system could perfectly well master language but be unable to show any coherent reasoning. Imagine, for example, an AI system that acted not like an LLM but like a fluent idiot producing grammatical sentences vaguely relevant to the topic but which otherwise made little sense. A conversation with such a system might look like this:\footnote{Try entering this prompt into a state-of-the art LLM and you will see a much more informative and cogent answer.}
\begin{quote}
\noindent
Me: Why do different US states have different birds as mascots?\\
AI: Birds build nests in many states.
\end{quote}
Such a system could converse in fluent idiomatic English but be incapable of correctly answering almost any questions about any subject. This system, would surely succeed admirably in language and its ability to write individually coherent sentences would be itself remarkable but it would not show noteworthy domain-general reasoning abilities.\footnote{\citet{MAHOWALD2024517} emphasize the difference between pure linguistic competence and the ability to reason. However, while they note substantial and systematic failures in LLMs reasoning capacity, they don't dispute the point that LLMs exhibit quite a lot of competence and common sense in reasoning.} LLMs by contrast, as we have seen, give extensive relevant answers, not just grammatically correct meaningful sentence on a similar topic. That, they do so, and do so without relying on  regurgitating training texts, is the demonstration of their general inferential abilities, not simple linguistic capacities.

Of course, some might claim that given the type of things LLMs are, despite the fact that they answer reasoning questions well, what they do cannot be \emph{reasoning}. According to this line of thought, LLMs lack the right mental states, intentions, or connections to the external world to be counted as reasoners.\footnote{These kind of skepticism about different mechanical reasoning devices has a long history, but perhaps reached its apogee with Searle's Chinese Room Argument \citep{searle1980minds}. The most prominent version of such skepticism applied to LLMs is their dismissal as ``stochastic parrots'' by \citet{bender-koller-2020-climbing,bender2021dangers}} My response to this is just that they are using ``reason'' in a different sense than I am.  Here, I am simply interested in whether a system can draw reasonable inferences (in the mechanical sense of spitting them out), not whether it shares other  features of human mental states that go along with this ability. Given the limited scope here, we can set aside the philosophical anguish over determining the ``real'' marks of the mental that dominated so much twentieth-century philosophy of mind.

That LLMs engage in strong general reasoning is the positive claim. The negative claim is  that contemporary AI systems that are not language based, don't show strong domain-general reasoning. Many of the recent  AI milestones  have not involved language. Facial recognition, automated driving, strategic game play are some of the many areas where neural networks have performed astoundingly without any linguistic training data. Nonetheless, non-language based AI systems  possess narrow, domain-specific capacities, making it a stretch to say they engage in domain-general reasoning. Or so I will argue.

That AI systems besides LLMs are not exhibiting strong general reasoning is a substantive empirical claim, not a conceptual truth. Reasoning across a variety of domains (such as the causal, the mathematical, and the agential) is certainly possible without language. Testing the reasoning abilities of non-linguistic minds is a standard part of at least two  fields in psychology: cognitive developmental and comparative psychology. Cognitive developmental psychologists use pre-linguistic infant's eye gaze and grasping to make inferences about their cognitive capacities \citep[see][for review]{spelke2022babies}. There is also a rich tradition of experimental and observational studies examining the reasoning capacities of all kinds of animals \citep[see][for a popular review]{de2016we}. Sophisticated animals like mammals and birds only survive by way of their capacity to make predictions and inferences about causal forces, locations, and other animals' mental states. That animals and infants can, without words, reveal reasoning capacities across many domains is not under debate, the only question is just how strong these capacities  are.

My claim is that, with the exception of LLMs, there is little evidence that today's AI systems engage in such strong domain-general reasoning. There are, I contend, no non-language-based AI systems that can reason well about causation, agency, as well as basic physics. Negative claims are harder to make than positive claims. We tend, rightly, to focus on the successes of AI systems rather than absence of skill, the lack of general reasoning in non-language based systems is not easy to see. Nonetheless progress in AI has not universal or uniform: we need to look at what AI systems \emph{cannot} do as well as what they \emph{can}. In explicit tasks of general non-verbal reasoning, AI systems despite all their successes still can lag behind humans \citep{Zerroug:2022ty}.

We can also draw inferences about the state of non-verbal reasoning in AI just by looking at what kinds of systems are available today. To take one example where the lack of general reasoning skills seems evident, consider systems that have been designed to play video games or strategic board games like go \citep{DBLP:journals/corr/MnihKSGAWR13,doi:10.1126/science.aar6404}. These systems only exhibit powerful play in video or board games after very extensive practice play, generally orders of magnitude more than human players require. That is, to play a particular video game at a high standard an AI system requires hundreds of hours of play on that particular game. Humans require much less time, presumably because our general world knowledge and reasoning  allow us to quickly hit on reasonable playing strategies \citep{Lake2017-LAKBMT}. The point is not that machines require more training than we do in general. LLMs are also trained on more linguistic data than humans are by several orders of magnitude \citep{frank2023bridging}. The point is rather about the skills achieved after training. No AI system, however much training it has, possesses the capacity to get up to speed on \emph{new} video game at the rate humans do. Such a capacity, no matter how much training it required, would reflect strong human-like  reasoning abilities. No AI system has this capacity.

Video game play illustrates the  point above that language-use is not  necessary to exhibit strong reasoning abilities across a number of domains. An AI system that could learn to play almost any existing video game up to a human-standard with human-level amounts of practice, would be a good candidate for an AI systems showing strong domain general reasoning. Playing a variety of video games well, after all, would seem to require reasoning about agents, intuitive physics, and tools. Such fast learning game-playing systems do not exist, despite video-game play being a major area of commercial research in AI.

The successes of systems for image-recognition, speech-transcription, automated driving, and image generation are remarkable. Nonetheless, many easily definable tasks are still beyond the capacities of even the best AI systems.\footnote{Here I am following from points made by Yann LeCunn's in various 2024 lecture slides. For example in, his \href{https://drive.google.com/drive/folders/0BxKBnD5y2M8NUXhZaXBCNXE4QlE?resourcekey=0-WtYv0wV-8DFNsFWfRUcpsw}{Harvard Ding-Shum Lecture} he writes, ``Generative prediction only works for text and other discrete modalities''.} Consider, for example, a system that tries to do for video something like what LLMs do for text. This could be called a video prediction or continuation system. Such a system is given the start of a video and must generate a ``plausible'' continuation of the video. Creating such system is an active area of research in AI \citep{oprea2020review}. To create plausible continuations of videos requires causal and agential reasoning: you need to leverage such domains of knowledge to predict how objects and agents will behave.  There is no equivalent to GPT-4 for video prediction today---and indeed the most impressive video generation AI's like the text-to-video system SORA by OpenAI relies on an LLM for its operation. There are no neural networks that learn intuitive physics, causal reasoning, and how agents behave just by being trained on videos, despite the potential availability of huge amounts of video online. In AI today these general reasoning skills are  unlocked only by using language in the training.\footnote{One problem here is it's not entirely clear what the training regime could be used to teach a system to generate plausible continuation videos.}

Let me be  clear that I am making a falsifiable empirical claim. If a video generation/prediction system that does not use an LLM (or any  language-based training) can systematically generate the continuations of videos in a way that exhibits strong causal, agential and numerical reasoning, then the claim I making is falsified. If game playing systems has human-level capacities to get up to speed on new video and board games at human-like rates, then my claim will be falsified. Indeed if any non-language-based AI systems performs a task that would seem to require strong general reasoning skills, than my claim is false.
If we simply assume that deep learning is going to conquer every problem we can throw at it: video prediction, fully automated driving etc, we can draw no conclusions from current successes except that they will be followed by others. While this is exactly the conclusion many draw, such wild AI optimism---not only about what AI will accomplish, but what it will accomplish without using a human language---seems to me unwarranted.

\section{Natural language makes inference tractable}\label{utilityoflanguage}

My thesis is that LLMs alone succeed at general reasoning because  natural language makes general reasoning computationally tractable.

By definition, a problem can only be solved by an AI system if it is made computationally tractable. When you look at successes in AI you can often get a sense of what features of a problem make the tractable. Some of the earliest triumphs of neural networks were in strategic board games like backgammon, chess and go \citep{tesauro1994td,silver2016mastering,doi:10.1126/science.aar6404}. Playing a board game requires choosing at each stage among a finite small number of moves. Moreover each state of the game, determined by the board position, can be compactly and discreetly represented. The  successes of neural networks using reinforcement learning at playing these games shows that good play at these games is computationally tractable. That these problems were tractable is widely attributed to the discreet, digital nature of the task and the clear reward signal (i.e. winning or losing).  Other problems, like good video prediction or designing a robot that can make a cup of coffee \citep{enwiki:1235093491} in a range of typical domestic settings have not yet proven to be tractable for neural networks. These tasks notably lack the discreet representation we give to board games.

What we can see with LLMs is that by encoding problems of general reasoning in natural language,  a  discreet  and efficient medium of representation, general inference becomes computationally tractable for neural networks. Non-linguistic tasks that might require general reasoning, like video prediction or learning general strategic game play, may not be tractable without the aid of a language-like representation system. Language makes the task of inference and prediction tractable by abstraction: linguistic representation can encode exactly the relevant information without including too much irrelevant detail.

The same information can be encoded by different media. The fact that someone stole my locked bicycle from outside my office can be conveyed by showing a video of my bike being stolen by someone or it can be conveyed by the message ``your bicycle has been stolen''.\footnote{Of course for these messages to work, they have to be given in the right context. Somehow the video must be presented as factual and the time it was shot made clear. Likewise linguistic message must come from a speaker seen as honest and knowledgeable.} The video, of course, will give much more detail about the incident, including the look of the bike, the method used to break the lock, the appearance of the thief, etc. But the video, even with modern compression methods, will also take more data to convey. The words ``your bicycle has been stolen'' takes about 192 bits to store on a computer with ACII encoding, a 15 second video of at the quality of a non-HD video call would take about 12,000,000 bits.\footnote{This is assuming 100 kilobytes per second for video, a low estimate for reasonable quality.} For the amount of data you need for that video, you could have sent the entirety of Tolstoy's \emph{War and Peace}. Of course, video takes more data because it conveys  more information than text: My point is that if all you want to convey is that the bike was stolen, text is massively more efficient. Using realistic video you can't easily abstract from irrelevant detail.\footnote{Of course, you can get data to be more abstract and data-efficient. Using realistic video blurring out background, switching to black and white, and reducing pixel count are all tricks to use less data. But it's hard to see how to get to the data-leanness of natural language while conveying content with moving images.} This is the data-efficiency of natural language: it is a representational format that allows us to quickly convey the kinds of things we are interested in.

Different forms of data-efficiency can be found in different ways of encoding language itself. Recording a series of words using sound files takes orders of magnitude more data than using text. Storing words in text is more efficient because it eliminates  inessential information, such as what words actual sound like. While text encoding of language is a human invention, our minds---even pre-literate ones---must also use compact encodings of language to store linguistic information. 

Text encodes language efficiently. But language itself, likewise, encodes information about the world (i.e.\ non-linguistic information) efficiently. By allowing a sharp focus on the information we care about, language is able to serve as an efficient medium for storing and reasoning about the things we care about. The symbolic abstraction of language, is doubtless necessary for this. By not using, for example, a visual medium, much detail can be eliminated. But symbolic abstraction alone is not enough: chess notations and number systems are symbolic abstractions that are not versatile enough to support domain general reasoning.  Language combines date-efficiency and domain generality.

To see why this data-efficiency matters for inference, we need to expand on what the computational problem of prediction and inference is. Inference is the  process of drawing conclusions from a body of information. From the fact that it is raining and that you are outside with no umbrella or hat, you can \emph{infer} that your head will get wet. Prediction is a species of inference: inferring from what has happened up to a certain point of time what will happen after that point. Much of AI, and more broadly machine learning, is dedicated to inference and prediction.  That inference is such a broad category is why it's not a truism to say that only LLMs perform general inference. A robot that successfully makes a cup of coffee in a wide range of kitchens that it has not seen in its training will likely be demonstrating inferential powers.  

Of course, there are different forms of inference such as deductive (or logical) inferences and inductive inference.  Any of these forms of inference can be used to support prediction as well, though, of course, many, following Hume's classic discussion, think no prediction can be purely deductive.\footnote{But that doesn't mean deductive inference can't support prediction. A physicist can derive predictions using mathematical deduction, even if the assumptions behind the mathematics are only supported inductively. (Apologies for the Reasoning 101.)} Classical AI is often employed for deductive reasoning. Deductive reasoning is, by contrast, the weak point of connectionist systems. If you want  a calculator, training a neural network is an extremely inefficient way to get one. 

But for the vast realm of non-deductive inference, the Great AI Experiment shows the power of connectionist approaches. Next-token prediction, predicting how texts continue, can be seen as a very basic computational model of all forms of inference within the medium of language. To put  it simply, we can think of an inference problem as one about what follows (deductively or in some other sense) from what.  A system that takes some sentences expressing information and returns more sentences expressing information can be seen as performing inference. The key question is how good the inferences are from a given system.\footnote{Of course, merely continuing a sentence is not naturally thought of as a form of inference. So only some next-token prediction can be viewed as inference.}

What's particularly surprising about LLMs is not the fact that they make some form of inference. That only requires them to formulate coherent English sentences  that convey some information. What is genuinely surprising rather the \emph{quality} and \emph{range} of the inferences.\footnote{A deductive inference machine, which can be designed with classical AI, may have high quality inferences, but in only performing deductive inference the range of inferences will be limited.} Indeed, in their ability to reason generally around a number of topics they seem to have either solved or made irrelevant the notorious frame problem in AI \citep{sep-frame-problem}.

When we see what the problem of inference is, we can see why the economy of language is critical for it. Predicting video is massively harder than predicting text because there are so many more possibilities. Predicting even the next word in a text is a difficult problem, but it is massively simpler than predicting the next frame in a video. There are  orders of magnitude fewer choices of words than  possible frames.\footnote{The problem of prediction is not really one about the sheer quantity of possibilities, but about their probability distribution. In the crucial information theoretic framework \citep{ShannonCommunication}, the more regularities there are to be exploited the less information there is to be encoded on average. Language in addition to being discreet is also extremely predictable.} Prediction and inference via language is simply computationally more tractable because language allows us to efficiently communicate exactly what we care about and what is relevant for prediction. That some reasonable degree of general inference  present in AI today may only be possible because of the efficiency of language. This provides a plausible explanation of why the only AI systems that succeed at general inference are the ones, in Dennett's terminology, that have had language added to them. Natural language provides a format that makes inference tractable.\footnote{Dennett himself doesn't relate, as far as I know, the usefulness of natural language in thought to its compression.   He does, though, suggest that we perform inference on linguistic representations:
\begin{quote}
 Once we have created labels, and the habit of ``attaching'' them to experienced circumstances, we have created a new class of objects that can themselves become the objects of all the pattern-recognition machinery, association-building machinery, and so forth. \citep{dennett1993learning}
\end{quote} Dennett here seems to see the advantage of language as allowing us to process what is stored in it repeatedly and at leisure not the fact that the storage itself so economical as to facilitate inference. I am not quite clear why Dennett dismisses the possibility that non-linguistic representations might serve these same purposes equally well.}

Let me contrast this answer about what language does for thought with a different standard explanation of the utility of language. This is the expressive power of language. Through language we can express almost any thought: at least the kind of thoughts that we can dream of. Natural language, in this sense, contrasts itself with, say, a chess or musical notation which can only be used to code the course of a game of chess or the sequence, timing, and intensity of notes in a piece of music. 

However, the kind of flexibility that comes with expressive power is necessary but not sufficient for general inferential abilities. Most neural networks are extraordinarily flexible: in mathematical terms a large enough neural network can approximate any well-behaved function. Hook such a system up to inputs and outputs and there is clearly a huge range of theoretical potential for general inference.\footnote{Similar remarks can be made about classical computers,  but there no well established  effective general learning mechanisms for classical AI, so it's much less clear to how to exploit such flexibility.} It is true that it's impossible to imagine a digital clock exhibiting general inference, but the same cannot be said of a huge artificial neural network given appropriate training. Computers do not need language in order to be flexible enough to deal with almost any problem they are presented with.

Of course, an AI system that isn't fed linguistic inputs can develop its own representation system through learning abstractions on its own.  Many deep neural networks with their hierarchical structures are designed to do exactly this: to abstract as they predict.\footnote{Predictive coding architectures like that proposed by \citet{friston2003learning} also combine abstraction with prediction in a method not far from  deep learning.} The successes of non-language-based networks on tasks like facial recognition, video game playing, automated driving all demonstrate the capacity of networks to abstract useful patterns out of the data and use these abstractions to drive their performance. But, so far at least, neural networks that do all the work of abstraction themselves are not capable of the level of general reasoning that LLMs show. Having language significantly reduces the computational task required to be solved by the network---making general inference possible. 

In other words, language might make general inference tractable by  reducing the problem an AI system needs to solve. To infer something useful about a visual scene a system might need to first abstract to the categories that matter for what we take good prediction to be, and then in that abstract format perform the inference.  When the problem is put entirely in terms of language that first stage of abstraction may be reduced if not eliminated altogether. That natural language encoding of information streams makes inference tractable, though is no a priori fact. Rather this is what we learn from the success of LLMs at general inference problems. We can then explain that success by the fact that natural language (unlike, say, video) abstractly represents inferentially significant properties in a compact way. 

I have argued here that language enables general inference because it provides a compact but (effectively) universal representation system. LLMs show that training an appropriately structured deep neural network with enough data encoded this way unlocks general inferential abilities. 

\section{Lessons for cognitive science}\label{lessons}

So far, the conclusions I have drawn solely concern the state of artificial intelligence. The massive artificial neural networks implementing LLMs and other contemporary AI systems have different architectures and training from those of human and animal brains. Nonetheless, the possibilities they show can help us understand our biological minds. 

Here I'm going to go through what I take to be some of the takeaway messages for cognitive science about the role of language in our thought.

\subsection{Language may also be shaped facilitate human inference}

Steven Pinker in \emph{The Language Instinct}, gives a spirited presentation of the idea that natural languages are optimized for efficient communication not efficient thought:
\begin{quote}Any particular thought in our head embraces a vast amount of information. But when it comes to communicating a thought to someone else, attention spans are short and mouths are slow. To get information into a listener's head in a reasonable amount of time, a speaker can encode only a fraction of the message into words and must count on the listener to fill in the rest. But inside a single head, the demands are different. Air time is not a limited resource: different parts of the brain are connected to one another directly with thick cables that can transfer huge amounts of information quickly. 
\end{quote}
Even among those who disagree with Pinker on his claims about the innateness of language his view that language is largely optimized for communication remain common.\footnote{Pinker's Harvard colleague Joseph Henrich, writing 22 years later, rejects Pinker's claim that the capacity to learn language is a largely a genetic adaptation, but he concurs that the shape of language is determined by communicative needs rather than anything else.
\begin{quote}
Converging lines of research from several fields now point to an answer: languages arise from long-term cumulative cultural evolution. Like other aspects of culture, including sophisticated technologies, rituals, and institutions, our repertoires of communicative tools--—including spoken languages—--have evolved via cultural transmission over generations to improve the efficiency and quality of communication, and to adapt to the details of local communication contexts, including physical environments and social norms (like taboos). Languages, then, are cultural adaptation for communication. \citep{henrich2016secret}
\end{quote}
\citet{gibson2019efficiency} gives a powerful expression of the view that language is optimized for  its use in communication.}

Pinker with his invocation of ``thick cables'' in the brain suggests that the degree of compression in language would have little value instrumental value for thought. The thought is intuitive: why bother to compress a 1 megabyte file if you are transferring it via a 12 megabyte per second connection? What Pinker and others don't consider is the potential utility of abstraction for the use of language as a medium of thought. If thoughts are not conveyed in a sufficiently compressed and abstract a manner, performing non-deductive inferences on them will simply be too difficult. The Great AI Experiment demonstrates the utility of natural language as an abstract medium for inference. This leaves open the possibility that the efficiency of language is shaped for this purpose rather than just communicative needs.

\subsection{Language as a driver  rather than  product of human cognitive powers}

That language is optimized for communication, rather than thought, naturally fits with the view that natural language does not play a major role in thought itself. \citet{fedorenko2024language} summarize this position as follows:
\begin{quote}
 We conclude that although the emergence of language has unquestionably transformed human culture, language does not appear to be a prerequisite for complex thought, including symbolic thought. Instead, language is a powerful tool for the transmission of cultural knowledge; it plausibly co-evolved with our thinking and reasoning capacities, and only reflects, rather than gives rise to, the signature sophistication of human cognition.
\end{quote}
On this view, the primary effect of language is through its communicative ability, including the capacity to pass down growing cultural knowledge through generations. Linguistic representation, then, reflects rather than determines the shape of our thought.

The question about the role of language in thought relates to    broader question about human minds: what explains the vast gulf between human and animal minds, Is language an  effect of this difference or a major driver of it? As we have seen, many cognitive scientists like Pinker and \citeauthor{fedorenko2024language} downplay the significance of language in the development of human thought. 
Many other prominent cognitive scientists see language not as the original driver of human cognitive power but a consequence of it. For example, Michael Tomasello argues that shared intentionality rather than language is the driver of human specialness \citep[e.g.\ ][]{tomasello2009cultural,tomasello2005understanding}. Taking a different tact, \citet{dehaene2022symbols}  contends that pre-linguistic symbolic reasoning skills is what gives humans a major cognitive advantage. 

The other position: that language is a central driver of our peculiar cognitive powers, is also common. \citet{Carruthers2002-CARTCF-2,CLARK2006370,DennettKinds,spelke2003makes} and \citet{lupyan2016language} are a few examples of prominent cognitive scientists who point to language itself as a key, if not \emph{the} key, element in human cognitive specialness.\footnote{Of course as some have pointed out there  are probably more than  \emph{one} important factors explaining human cognitive specialness \citep{SheaConcepts}.}

There are a variety of reasons to think equipping minds with a language will change them. In \textit{The Descent of Man}, \citet{DarwinDescent} contended that  it is language use that enables us to engage in long trains of thought: 
\begin{quote} A long and complex train of thought can no more be carried on without the aid of words, whether spoken or silent, than a long calculation without the use of figures or algebra. (Vol.\ 1, ch. 2)
\end{quote} 
Later theorists have suggested even more foundational  effects of language on thought. On one school of thought, it is natural language which allows us to combine representations from different domains \citep{Carruthers2002-CARTCF-2,spelke2003makes}. \citet{spelke2003makes} writes: \begin{quote}
Natural languages provide humans with a unique system for combining flexibly the representations they share with other animals. The resulting combinations are unique to humans and account for unique aspects of human intelligence.
\end{quote} An even more radical standpoint is taken by Dennett who suggests that all conceptual thought depends on language: ``Concepts are things in our world because we have language'' \citeyearpar{dennett1993learning}.\footnote{While Dennett's is an extreme position in cognitive science, it has at times resonated with mainstream views in philosophy \citep[e.g.][]{Dummett1993-DUMOOA}.}

What light, if any, do the results of the Great AI Experiment shed on this debate? One thing they highlight is the \emph{possibility} that learning a language, along with much information conveyed in that language, to be cognitively transformative. The surprising result that simply training a system on masses of linguistic data would unlock strong domain general inferential capacities makes the case easier for those who think the acquisition of language has transformed our minds (and done so not merely via its communicative powers). Moreover, the considerations here suggest a specific route by which language can have a transformative effect: namely making  domain-general inference possible. In doing this, language may give humans a remarkable cognitive advantage over other creatures. 

This particular take on the effect of  language on thought,
 is a moderate one, not widely explored in the literature. Language, in this view, is not necessary to unlock ``higher'' cognitive abilities. Nor does  language need to be the only way to connect thoughts across different cognitive domains \citep{Carruthers2002-CARTCF-2}. The position is compatible, for instance, with \citeauthor{fedorenko2024language}'s \citeyear{fedorenko2024language} claim that ``language is not necessary for any tested form of thought.'' What language does is make some forms of inference much more tractable, increasing our facility with it. But one lesson of a century of AI is computational tractability matters: as the biggest driver of progress in AI has often been increased computational powers.\footnote{For example, the recent success of connectionist networks has only been possible with the development of ultra-fast and specially engineered GPUs to run them on. In classical AI, while super-human chess play was unlocked in the 1990s, Go-playing systems were far out of reach because of the increased computational complexity of the game.}  

 Interesting, LLMs suggest a nuance as well as to how one transforms a mind with language. On a view like Spelke's, Carruther's and Dennett's the mere possession of a new representational capacity, language, is transformative. For LLMs, however, it is not the language alone that unlocks cognitive benefits, it is the exposure to large amounts of linguistically encoded information. For LLMs, learning a language  is not separable from learning large amounts of information about the world encoded in language. Of course, the same is true for children.  Thus, it may not be the bare learning of a linguistic representational system that effects a cognitive transformation. What is learned is not just the propositional information encoded in each sentences, but also probabilistic facts about which words likely follow which.\footnote{But in children, unlike in LLMs, of course, learning a language also unlocks the potential to encode sensory data in linguistic form, creating a new sources of data for inferential inference.} 

I was careful to note above that what LLMs show is the \emph{possibility} of the transformative effect of language. LLMs are different  from human minds: they contain vastly less innate structure and organization and they are trained on orders of magnitude more linguistic data than any of us will encounter in a lifetime \citep{frank2023bridging}. They also lack all of the non-linguistic capacities of human minds. We know what LLMs are doing is not simply regurgitating texts or anything in that neighborhood, but whatever they are doing is doubtless profoundly different from what we do when we speak and write. So we must be cautious of drawing inferences from the success of LLMs about the role of language in human thought.

For AI systems the abstractions of natural language enable inferential powers unreachable by other systems. LLMs receive these abstractions, in a sense, pre-made in  the linguistic data that they are trained on. In this way, LLMs differ profoundly from other AI systems, like the convolutional neural networks used for processing visual information, which must, through training, develop on their own some of the abstract structure of the visual world.\footnote{Of course, the hierarchical structure of an untrained convolutional neural network already contains considerable structure, but it is only through training that skills like boundary recognition necessary for parsing visual scenes are developed \citep{fukushima1982neocognitron,lecun1998gradient ,SmolenskyMcCoy}.} 

Unlike LLMs, us humans doubtless learn many of the abstractions made by language without the \emph{aid} of language. The human parsing of the world into one of objects, agents, and causation, is an abstraction that, to a significant degree, natural language must reflect rather than create. After all, non-linguistic creatures also show inferential facility with objects, agents and causes, and there is considerable evidence that much of human facility with such basic concepts has an innate basis \citep[see][for a comprehensive review]{spelke2022babies}. Unlike LLMs, humans have access to huge amounts of non-linguistic data for training both in their own lifetimes and through evolutionary learning processes. The empirical question is really one of \emph{the extent} to which the abstractions of natural language are reflective of pre-linguistic cognitive processing and to what extent they come from language. By showing a route by which strong inferential abilities can appear through the use of language alone, LLMs make more plausible the view that many of the abstractions of natural language are not merely reflective of human cognitive capacities but directly encoded in language itself.

As with many important debates in cognitive science, the key issue seems to be one of degree not kind. There is direct evidence that some of our cognitive skills are achieved through a combination of pre-linguistic and linguistic cognition. Decades of research on the acquisition of mathematical knowledge has unveiled a key role for symbolic counting systems in unlocking some quite basic mathematical cognition, including the understanding of precise cardinalities above 3 \citep[e.g.][]{DehaeneNumber,CareyDaedalus}. It seems that a symbolic system is needed for the human mind to understand what it means to have a group of four or five animals. But other mathematical concepts like the cardinalities less than four and the basic concept of a ``a greater number'' do not appear to require language. Research on numerical cognition seems to show that symbolic systems both encode preexisting cognitive abstractions and help support the acquisition of new ones.

The empirical studies of basic numerical concepts along with the abilities of LLMs together make a strong case for the \emph{possibility} of profound affects of language learning on cognition. The notion that this is more than a mere possibility by the \emph{lack of evidence} for any innate natural-language like language of though bolstering human cognition. On a Fodorian \citeyearpar{FodorLOT} view natural language will simply be a means of communicating thoughts which are couched in a pre-linguistic language-like representation system, a general language of thought. This version of the language of thought hypothesis, on which there is a general LOT distinct from and prior to natural language, naturally relegates language to a minor role in guiding cognition.

The problem with this extreme LOT hypothesis is that there is, in fact, little evidence for it. Even in their  powerful recent defense of the LOT hypothesis, \citet{Quilty-Dunn2023-QUITBG} make no case for one general, natural-language like LOT but rather suggest that smaller, different LOTs may operate behind different mental operations \citep[see also][]{modularity}. Even cognitive developmental scientists with great regard for Fodor, like  \citet{spelke2003makes} and  \citet{CareyDaedalus,Carey_2023} argue against an innate general LOT with anything like the expressive power of natural language. Indeed it remains an open question whether, without language, we have access to a concept like  natural language disjunction \citep{Carey_2023}. Arguments for a general LOT are also undermined by the very success of LLMs. LLMs are a possibility proof that a system can learn to reason with natural language without first possessing an LOT, a possibility Fodor \citeyearpar{FodorLOT} dismissed. 

If there is no general language-like LOT prior to natural language, then   natural language may fill this cognitive void. That is, without a pre-existing system with scope and representational capacity of natural language, it is plausible that once one learns a language it will play a major role in cognition.\footnote{However, there are other possibilities. In recent work, for example, \citet{dehaene2022symbols} have argued for a pre-linguistic penchant for compressed symbolic representation.  If they are right it might be this symbolic ability that explains both our facility with language and with general inference.}

\section{Conclusion}

Equipping AI systems with langauge, by training them on masses of linguistic data, unlocks inferential powers that outperform all other types of AI. I argued here that is the abstract nature of natural langauges that makes this possible: they allow us to convey the kinds of things we care about with very little data. This means an AI system trained on language can effectively get much more useful training--for inference--in the same amount of time as an AI system trained on more naturalistic data. Turning to our own minds, it may well be the efficiency of linguistic representation has a profound influence on our own mental development.

\bibliography{danbib}

\end{document}